\documentclass{article}

\usepackage{arxiv}
\usepackage[utf8]{inputenc} 
\usepackage[T1]{fontenc}    
\usepackage{url}            
\usepackage{amsfonts}       
\usepackage{graphicx}
\usepackage{array}
\usepackage{multirow}
\usepackage{amsmath}

\newcommand\Tstrut{\rule{0pt}{2.6ex}}         

\title{Artificial Intelligence Prediction of Stock Prices using Social Media}

\author{
  Kavyashree Ranawat\\
  Durham University School of Engineering and Computing Sciences\\
  Durham University\\
  Lower Mountjoy, South Rd, Durham DH1 3LE, United Kingdom\\
  \texttt{kavya.ranawat@gmail.com} \\
   \And
 Stefano Giani \\
  Durham University School of Engineering and Computing Sciences\\
  Durham University\\
  Lower Mountjoy, South Rd, Durham DH1 3LE, United Kingdom\\
  \texttt{stefano.giani@durham.ac.uk} \\
}

\begin{document}
\maketitle

\begin{abstract}
The primary objective of this work is to develop a Neural Network based on LSTM to predict stock market movements using tweets. Word embeddings, used in the LSTM network, are initialised using Stanford's GloVe embeddings, pretrained specifically on 2 billion tweets. To overcome the limited size of the dataset, an augmentation strategy is proposed to split each input sequence into 150 subsets. To achieve further improvements in the original configuration, hyperparameter optimisation is performed. The effects of variation in hyperparameters such as dropout rate, batch size, and LSTM hidden state output size are assessed individually. Furthermore, an exhaustive set of parameter combinations is examined to determine the optimal model configuration. The best performance on the validation dataset is achieved by hyperparameter combination 0.4,8,100 for the dropout, batch size, and hidden units respectively. The final testing accuracy of the model is 76.14\%. 
\end{abstract}


\section*{Introduction}
Twitter is a microblogging and social media platform that allows users to communicate via short messages (280 characters) known as tweets \cite{one,two,three}. It enables millions of users to express their opinions on a daily basis on a variety of different topics ranging from reviews on products and services to users' political and religious views, making Twitter a potent tool for gauging public sentiment \cite{four}. Thus, it manifestly follows that twitter data can be regarded as a corpus, forming the basis on which predictions can be made, and researchers have indeed exploited this fact to seek trends by performing numerous and varied analyses.
 
A characteristic feature of the stock market is volatility and there is no general equation describing the prediction of stock prices, which is a complex function of a range of different factors. The methods of stock market prediction can be broadly classified into Technical Analysis and Fundamental Analysis \cite{five}. The latter involves the consideration of macroeconomic factors as well as industry specific news and events to guide investment strategies \cite{five}. The analysis of public sentiment via tweets performed in this project can be regarded as an aspect of Fundamental Analysis. Although the prediction of stock prices is highly nuanced, the Efficient Market Hypothesis (EMH), propounded by Eugene Farma in the 1960’s, suggested a relation between public opinion and stock prices \cite{six}. The semi-strong form of the EMH implies that current events and new public information have a significant bearing on market trends \cite{one,six}. This view is supported by Nosfinger who draws upon evidence from several studies in the field of Behavioural Finance to reinforce that changes in aggregate stock price as well as the high degree of market volatility can be, in part, attributed to public emotion \cite{seven}. Numerous studies have been successful in unveiling and proving the perceived existence of a relationship between public mood gathered from social media and stock market trends \cite{seven,eight,nine,ten, eleven,twelve,thirteen,fourteen,fifteen}. 

The vast majority of ML (Machine Learning) techniques applied in this sector have integrated the characteristics of NLP (Natural Language Processing) to extract and quantify the sentiment of public opinion expressed via social media. SVM (Support Vector Machines) \cite{eight}, Random Forest \cite{nine}, and KNN (K-Nearest Neighbour) \cite{eleven} classifiers have yielded impressive classification accuracies in this application. The only caveat is that most studies consider the compound effect of historic prices and public sentiment, thereby discounting the exclusive impact of sentiment.

Some studies have gone beyond the classic ML approach, employing deep learning methods such as ANNs (Artificial Neural Networks) in one form or another for the purposes of making predictions \cite{twelve,thirteen,fourteen,fifteen, sixteen, seventeen}. Bollen et al have established that collective public mood is predictive of DJIA (Dow Jones Industrial Average) closing values by making use of Granger Causality Analysis and SOFNN (Self-Organized Fuzzy Neural Networks). Many researchers have built on this work and others have explored alternate deep learning models such as MLP (Multi Layer Perceptron), CNN (Convolutional Neural Network) + LSTM (Long Short Term Memory) for market prediction.

The primary focus of this work was on the development of a variant of RNN (Recursive Neural Network), known as LSTM, capable of predicting short-term price movements. Owing to the volatile and unpredictable nature of the stock market, it is plausible that the relationship between the societal mood and economic indicators perhaps is more complex and nuanced than linear. Deep learning methods are felicitous for this application in that hidden layers can exploit the inherent relational complexity and can potentially extract these implicit relationships. It is for this reason that an LSTM structure was selected as the principal model in this work. The popularity of RNNs in NLP and stock prediction tasks is attributed to the fact that they consider the temporal effect of events which is a significant advantage over other NNs (Neural Networks). With the aid of a popular sentiment analysis tool, known as VADER, the degree of correlation between the sentiment expressed via tweets and stock price direction was also investigated for the purposes of comparison with the results from the LSTM architecture.

\section{VADER Implementation}\label{theory}

VADER is a state of the art technique employed by researchers in sentiment analysis tasks. One aspect of this work involves using VADER to explore the degree of correlation between public opinion and sentiment expressed via twitter and stock market direction.  As aforementioned, VADER is a gold standard lexicon and rule-based tool for sentiment analysis \cite{eighteen,nineteen}. Developed and empirically validated by Hutto and Gilbert, the VADER lexicon is characteristically attuned to text segments in the social media domain \cite{twenty,twentyone}. Unlike other lexicon approaches, VADER takes into account that microblog text often contains slang, emoticons, and abbreviated text \cite{twentyone}. It not only provides the semantic orientation of words but also quantifies sentiment intensity by considering generalisable heuristics such as word order, capitalisation, degree modifiers etc \cite{twentyone}. In this application, VADER was used to generate the polarity scores of tweets, including a compound score (normalised between 1 and -1) which reflects the combined effect of the degree of positivity, negativity, and neutrality expressed in a tweet. 

\subsection{Experimental Procedure}  
The aim is to investigate the correlation between two variables; VADER scores and stock market trajectory. Firstly, tweets containing the APPLE stock ticker symbol were cleaned, using the algorithm described in the next section. Stock data associated with APPLE, among the Big Four technology companies, is deemed as a suitable choice upon which to perform analysis for several reasons; a detailed rationale is provided in a subsequent section.  The compound score for each tweet was generated using VADER. Subsequently, the average of the scores for all tweets in a single day was taken. To obtain values for the second variable i.e. stock market movement, the direction of stock price movement was quantified. If the next-day close price of the security is greater than that of the current day, the value for that day is defined as 1, else it is defined as 0. 

After generating the values for both variables, a special case of the Pearson coefficient, known as Point biserial correlation coefficient, was applied on the data to determine the correlation. This metric is commonly used when one variable is continuous and the other is categorical, as is the case in this application, where the VADER scores are continuous whereas the stock price change data is dichotomous (binary) \cite{twentytwo}. The biserial correlation method, however, requires the continuous variable to be normally distributed \cite{twentytwo}. Therefore, the distribution of the VADER scores was plotted and a roughly normal distribution was obtained (as shown in figure 1), allowing the use of the biserial correlation method.

\begin{figure}
\centering
\includegraphics[height=2in,width=0.5\columnwidth]{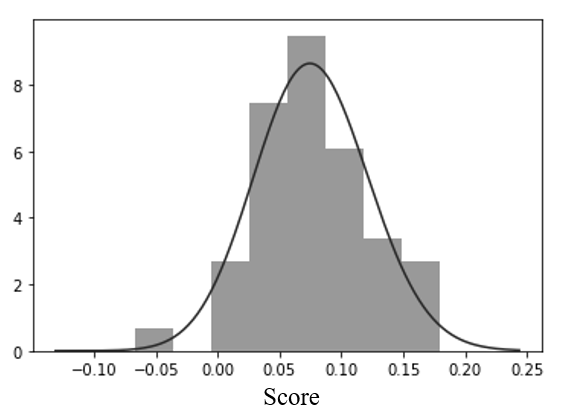}
\caption{Distribution of VADER scores (continuous variable).  }
\end{figure}

\subsection{Further Modifications} 

Modifications were made to the stock price change data by redefining the classification strategy. Previously, it was based on a delay of 1 day. It is plausible, however, that the effect of information and opinions presented in tweets may take longer to manifest and reflect in the asset price. To explore this theory, a delay size of days in the range [1,7] was taken. For example, a delay size of 2 indicates that if the asset price at the close of the trading day 2 days hence is higher, it is classed as 1, else it is classed as 0. 

\begin{figure}[h]
\centering
\includegraphics[height=2.2in,width=0.5\columnwidth]{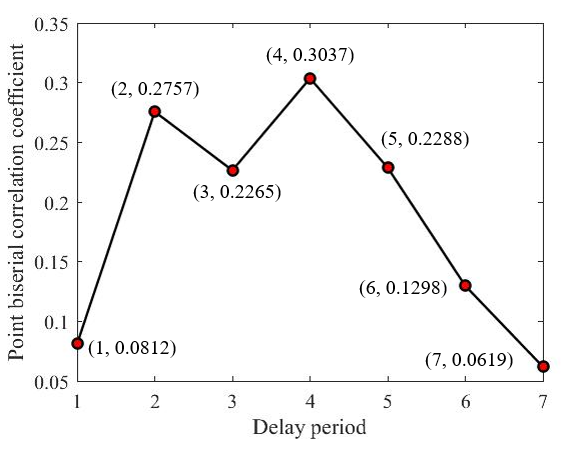}
\caption{Correlation coefficient values for different delay periods.  }
\end{figure}

Another configuration was assessed after making modifications to the VADER scores. If a particular tweet has more retweets than others i.e. it has been frequently shared by users, it could indicate that the information contained in that tweet has a notable influence on other users. These tweets could potentially play a greater role in impacting an incremental change within the stock market. By taking a simple average of the compound scores of tweets to arrive at a singular value for a specific day, it is not possible to gauge the contribution of important tweets. Thus, based on the number of retweets per tweet, a weighted average of the compound scores was taken to ensure accurate representation of all tweets. The correlation was then determined using the modified scores. 

\subsection{Results}
The correlation obtained for variable delay configurations is shown graphically in figure 2. It is evident that the correlation for the original configuration using a delay of 1 day is 0.0812 (or 8.12\%), suggesting a notably poor linear relationship between the sentiment expressed in tweets and market movement. Using delays higher than 1 generally tend to improve the degree of correlation, with a delay of 4 days yielding the highest correlation (30.37\%). This result supports the assumption that there exists a lag between the release of public opinion and its consequent reflection in stock price. A decreasing trend is observed as the delay size is increased beyond 4 days. A potential implication of this could be that the information contained on a particular day is irrelevant after large delay periods and no longer has any bearing on stock market values.

The retweets-based weighted average configuration results in a small negative correlation value of -8.87\%, which is in complete contrast to the original configuration. There is ambiguity in what the underlying cause of the generated results could be. There is a possibility that the inclusion of retweets has no impact in moulding future stock values. Alternatively, it could also be that the lack in the number of data points is masking the true contribution of the retweets. 

It can be concluded from the results that VADER is not able to present any strong association between public sentiment and market trajectory. The low correlation values could be attributed to the use of an insufficient number of samples for both variables. The final VADER scores are derived by taking an average of all compound scores, which is a simplistic and naive approach. The inability of the lexicon-based tool to identify any notable relationships between the variables could also be due to the inadequacies inherent in the development of VADER sentiment. However, as mentioned earlier, there is a high possibility that the relationship between the variables is not linear and is perhaps more complex and nuanced. This could be one of the reasons why using a metric such as correlation, used to assess strength of a linear relationship, is not able to detect any significant associations.

\section{Neural Network Model} \label{Neural Network Model}
ANNs, inspired by the behaviour of neurons in biological systems, are a dense interconnection of nodes or pre-processing units connected in layers, which have the ability to discover complex relationships between inputs and outputs \cite{twentythree}. There are many varying implementations of ANNs, which differ in terms of network architecture, properties and complexity. Considering the sequential nature of tweets, it is essential to use a network which considers the temporal effect of an input sequence. RNNs satisfy this criterion and are indeed suitable for application in tasks of this nature. However, the main drawback of RNNs is their inability in capturing long term dependencies in an input sequence i.e. the Vanishing Gradient problem \cite{twentyfour, twentyfive}. For example, when an input sentence is fed into the network, the error must be backpropagated through the network in order to update the weights. If the input is long, the gradients diminish exponentially during backpropagation, resulting in virtually no contribution from the state in earlier time steps. It is particularly problematic when using the sigmoid activation function as its derivative lies in the range [0, 0.25], resulting in highly diminished gradient values after repeated multiplication. A variant of the vanilla RNN network, the LSTM, can overcome this limitation \cite{twentyfour, twentyfive}, albeit not entirely, and is used in this application. Figure 3 shows a high level abstraction of the LSTM model architecture, consisting of one stacked LSTM layer. The input side shows the vectors, concatenated in the Embedding Layer, being input to the LSTM cells in the stacked layer. The output side depicts the hidden state outputs of the LSTM layer being taken as inputs by a sigmoid-activated node to make output predictions.

\begin{figure*}
\centering
\includegraphics[height=4in,width=1.03\textwidth]{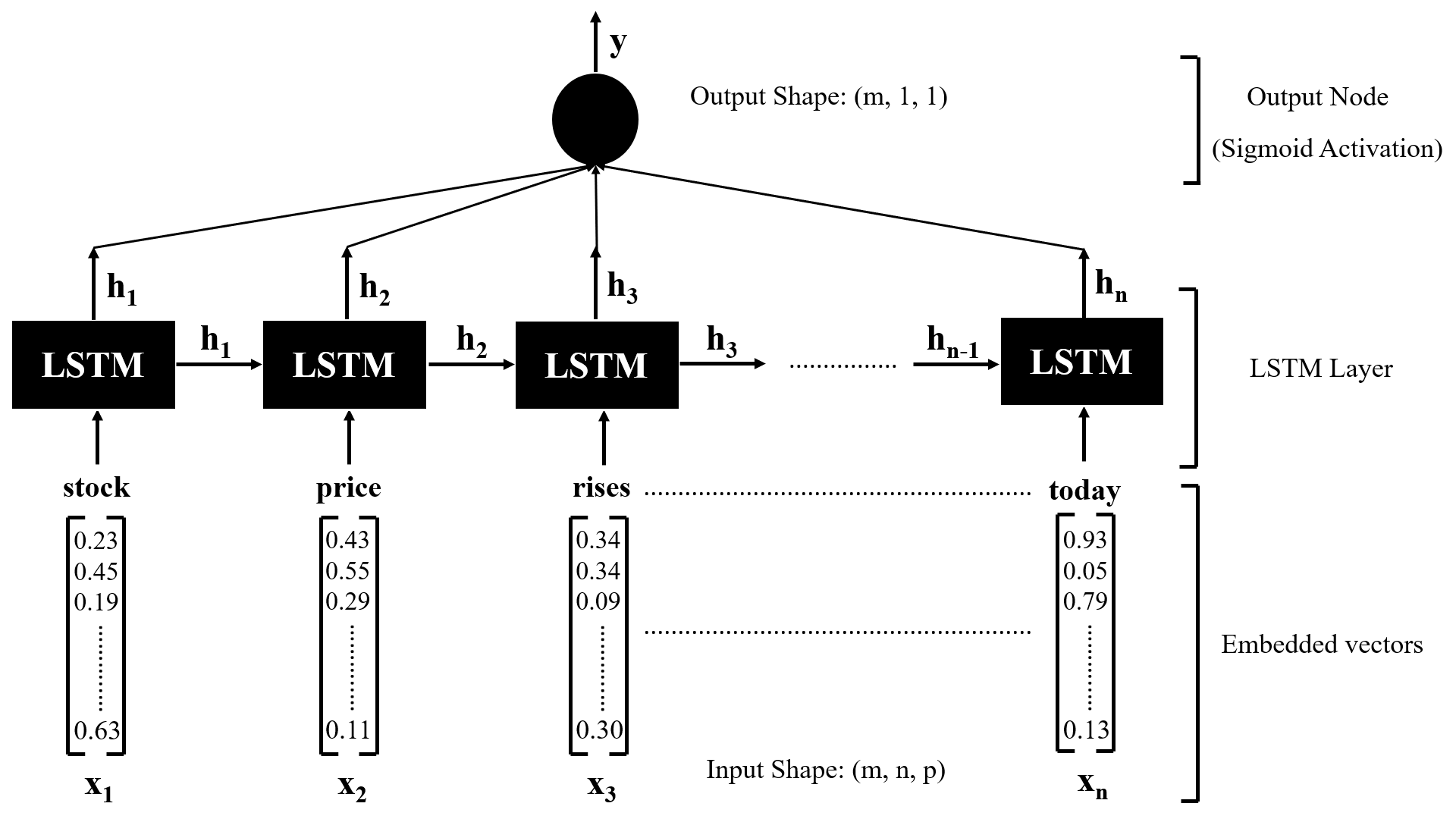}
\caption{General schematic of the LSTM Network. The input side shows the vectors, concatenated in the Embedding Layer, being input to the LSTM cells in the stacked layer. The output side depicts the hidden state outputs of the LSTM layer being taken as inputs by a sigmoid-activated node to make output predictions.}
\end{figure*}

\subsection{Preprocessing}
Raw tweets contain a great deal of noise which needs to be eliminated in order to extract relevant information from tweets and improve the predictive performance of the applied algorithm. Tweets contain twitter handles, URLs, numeric characters, and punctuation which do not contribute meaningfully to the analysis. A preprocessing algorithm was applied to remove these elements, convert words to lower case, and tokenize the words present in a tweet. The cleaned tweets were concatenated so as to form one input sequence for a given day. Ordinal or integer encoding was used to map all the words in the vocabulary to an integer value, resulting in an input vector $\{w_1, w_2, w_3,…,w_n\}$, where $w_{t}$ corresponds to a unique integer index representing a feature (or word) in the vocabulary. Although LSTMs can take inputs of variable length \cite{twentysix}, post-padding was applied as only vectors with homogeneous dimensionality can be used with the Keras Embedding Layer. For each input vector of length $t \in [1,n]$, $n-t$ zeros or dummy features must be appended, where $n$ is the length of the longest encoded vector. This produces vectors in an $m$-dimensional feature space, where $m$ is the total number of unique samples i.e. number of unique phrases or tweets fed to the network.
\subsection{Embedding}
NLP tasks for textual representation and feature extraction commonly use BoW (Bag of Words) models owing to their flexibility and simplicity. The traditional BoW has two variants: N-gram BoW and TF-IDF (Term Frequency-Inverse Document Frequency). The former reduces dimensionality of the feature set by extracting phrases comprising $N$ words while the latter considers the frequency of words whilst considering the effect of rare words. The main drawback is that BoW fails to take into account word order and context and results in sparse representations. This work explores the use of GloVe (Global Vectors for Word Representation) which has gained momentum in text classification problems \cite{twentyseven,twentyeight}. GloVe overcomes the sparsity problems associated with the BoW model by generating dense vector representations and projecting the vectors to a markedly lower dimensional space. It has the ability of capturing the semantic and syntactic relationships that are present between words, where words with a similar meaning are locally clustered in the vector space. Stanford’s GloVe embeddings, trained specifically on 2 billion tweets, were used to project each feature as a 200-dimensional vector \cite{twentyseven,twentyeight}. The weights of the feature vectors were initialised using the pre-trained embeddings but adjusted with the progression of training to improve classification performance. Each integer encoded feature, $w_t$, corresponds to an embedding vector, $\textbf{x}_t$, where $\textbf{x}_t \in \mathbb{R}^p$. Owing to the fact that each feature is represented as a 200-dimensional vector, $p=200$. A padded input feature vector is thus represented in the embedding layer as $\textbf{X} \in \mathbb{R}^{np}$, formed as a result of concatenation of $m$ vector embeddings. In figure 3, although the embedding layer is not explicitly shown, the vector embeddings which constitute it can be seen as inputs to the LSTM layer at the corresponding time steps.

\subsection{LSTM Layer}
The main distinction between a neuron in the vanilla RNN and an LSTM cell lies in the presence of a cell state vector, whose contents at each time state are maintained and modified via an LSTM gating mechanism \cite{twentynine, thirty, thirtyone}. The information flow in an LSTM memory cell is regulated by three primary gates viz. forget gate, input gate, and output gate \cite{twentynine, thirty}. Figure 4 shows the schematic of an LSTM cell, including the gating mechanisms used to achieve its functionality.

\begin{figure}
\centering
\includegraphics[height=1.8in,width=0.5\columnwidth]{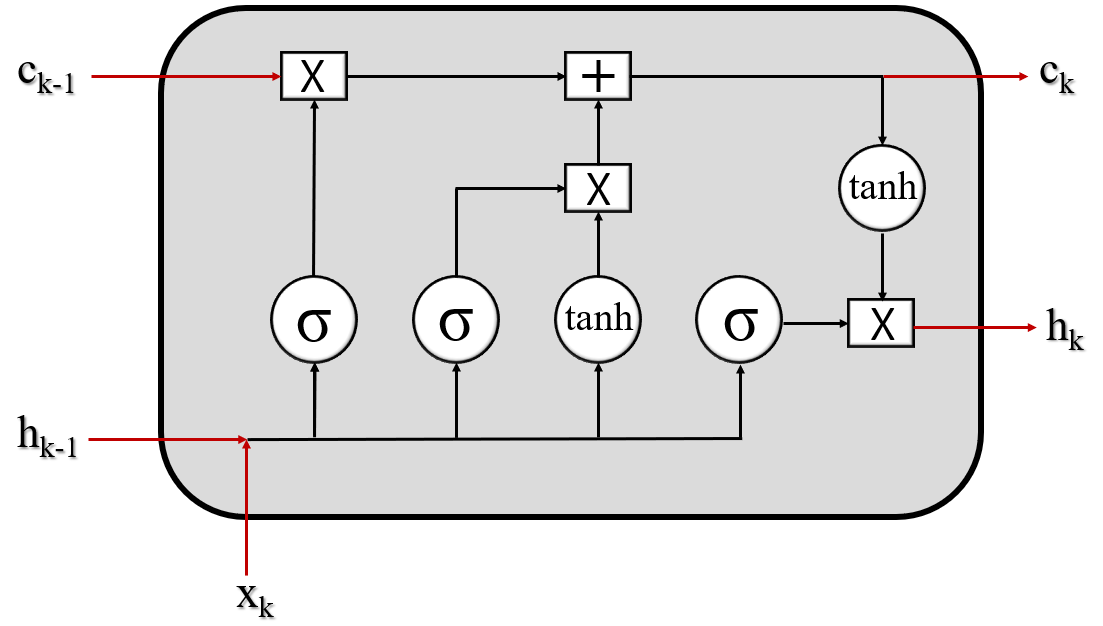}
\caption{Structure of an LSTM cell, showing the role of the primary gates and flow of information within the cell to form the current memory state and hidden output state. }
\end{figure}

At a certain time step, $k$, the vector embedding, $\textbf{x}_k$, along with the previous hidden output, $\textbf{h}_k$, will be used by the input and forget gates to update the internal state of the cell \cite{twentynine, thirtytwo}. The output gate combines the inputs and the current cell state, $\textbf{c}_k$, to determine the information to be carried over to the next cell in the repeating structure \cite{twentynine}. In this fashion, LSTMs can control the contribution of those words and word relationships that have a higher impact on prediction, whilst penalising those that are less significant. Equation 1 is a matrix representation \cite{thirtytwo} of the outputs of the gates. Equations 2 and 3 show the outputs of the cell, where the output vector is obtained by an element wise multiplication process \cite{twentynine, thirtytwo}.

\begin{equation}
\begin{pmatrix}
    i \\
    f \\
    o \\
    g  \\
    
\end{pmatrix} =
\begin{pmatrix}
    \sigma \\
    \sigma \\
    \sigma \\
    \tanh  \\
    
\end{pmatrix}  W 
\begin{pmatrix}
    x_k \\
    h_{k-1} 
    
\end{pmatrix}
\end{equation}

\begin{align}
c_k &= f \odot c_{k-1}+ i \odot g           \\
h_k &= o \odot \tanh c_{k}         
\end{align}

\noindent where $\textbf{i}, \textbf{f},$ and $\textbf{o}$ are the outputs of the input gate, forget gate, and output gate respectively and $\textbf{g}$ is the output of an additional gate which aids in updating cell memory. \textbf{W} is the weights matrix and $\sigma$ and $\tanh$ represent the sigmoid and $\tanh$ non linearities. Note that the system of equations in (1) also contains a bias term for each gate output. Dropout, a regularization method, is utilized to prevent the model from overfitting \cite{twentyfive}. Overfitting occurs when the model learns the statistical noise present in the dataset, capturing unnecessary complex relationships and thus, resulting in decreased generalisability. During training, it is possible for neighbouring neurons to become co-dependent, inhibiting the effectiveness of individual neurons. Dropout causes a proportion of the nodes or outputs in the layer to become inactive, thereby forcing the model to become more robust. This results in an increase in the network weights, which must be scaled by the dropout rate after completion of training.
\subsection{Output Layer}
A singular output node with a sigmoid activation function, presented in equation 4, was used for the purpose of classifying trend. 
\begin{equation}
\sigma(z)=\frac{1}{1+e^{-z}}
\end{equation}
\noindent where $z$ is the activation of the output node. The estimated probability returned by the node was compared against a threshold probability in order to perform binary classification. If the output probability for a given input sequence $\sigma(z) \geq 0.5$, the input was labelled as 1, predicting an increase in asset price for the following trading day. If the output probability did not exceed this threshold, the input was labelled as 0, indicating either no change or a decrease in next-day price. A binary cross entropy cost function,   $J_{bce}$, was used as given by equation 5 \cite{thirtythree}.
\begin{equation}
\begin{aligned}
J_{bce} &=-\frac{1}{m}\sum_{j=1}^{m}[y_j \times log(\sigma(z_j))\\
        &+(1-y_j) \times log(1-\sigma(z_j))]
\end{aligned}
\end{equation}
\noindent where $y_j$ is the $j^{th}$ target variable or the actual class label from a set of $m$ training samples. The cost or error function is representative of how accurately the model predicts target values, using a given set of network parameters. The main aim is to optimise or minimise the cost function, updating the weights and biases of connections in the network as a result. A mini-batch Stochastic Gradient was used during backpropagation to allow the model to converge to a global cost minimum \cite{twentyfive}. This optimum state represents a model configuration where the error between the actual and predicted values is at a minimum and the network can successfully detect patterns between word embeddings essential for classification. The learning rate determines the rate at which the tunable weights approach the global minimum and must be chosen judiciously. A very large value would risk overshooting the minimum and a learning rate that is too small will significantly delay convergence.

\section{Experimental Procedure}
Twitter data was obtained from \textit{followthehashtag} \cite{thirtyfour}, an online resource containing a readily available corpus of tweets. Approximately 167,000 tweets mentioning or associated with APPLE stocks were used for analysis. APPLE is among the companies currently dominating the technological sector and is regarded as a suitable choice upon which to base analysis. Owing to its popularity and the fact that it has the largest market capitalisation out of all NASDAQ 100 companies, it is fair to assume that twitter contains sufficient information relating to its stocks. The stock price data was sourced from Yahoo Finance \cite{thirtyfive}. The granularity of stock data considered is 1 day i.e. daily changes in stock price were computed to capture the essence of short term price fluctuation. The input tweets were labelled according to the scheme described previously. Concatenation of tweets leads to an aggregate of 48 input samples, corresponding to 48 unique days. Due to the limited number of samples, each  input was divided into 100 subsets, whilst keeping the labelling of the subsets consistent with that of the original day. Subsequently, 4800 input samples for the network were obtained. For the initial experimentation, a naive model configuration was used. This model forms the basis on which further improvements in performance can be achieved. The next section discusses the effects of using different network types, hyperparameter optimisation, and varying split values (for tuning the number of input samples) on model performance. For the initial configuration, consisting of 4800 samples, the training/testing/validation split was 70/20/10 i.e. training was performed on 3360 samples, testing was performed on 960 samples, and validation was performed on the remaining 480 samples. The validation set is used to configure the model so as to obtain the hyperparameters which give the best performance. The testing data is only used once after the network has been configured to give an unbiased evaluation of model performance. A single LSTM stacked layer was utilised with a dropout value of 0.2 and 100 hidden units (used for determining the dimension of the LSTM outputs). The gradients and weights are updated according to a batch size of 32.

The performance of the initial configuration is reported in the next section. The primary metric used to assess performance is accuracy. Another commonly used metric is the F1 score, which is the harmonic mean of the precision and recall \cite{thirtysix}. Precision refers to the percentage of instances correctly predicted as positive with respect to all instances classified as positive by the model, whereas recall refers to the percentage correctly classified as positive out of all positive classes \cite{thirtysix}. The F1 score is also calculated for varying implementations discussed in the next section . However, it is only needed when there is a greater cost associated with either the false positives or false negatives. As this is not applicable for this task and class distribution is even (as shown in figure 5), it is only computed to ensure consistency in results and thus not reported for all configurations.

\begin{figure}
\centering
\includegraphics[height=2.5in,width=0.6\columnwidth]{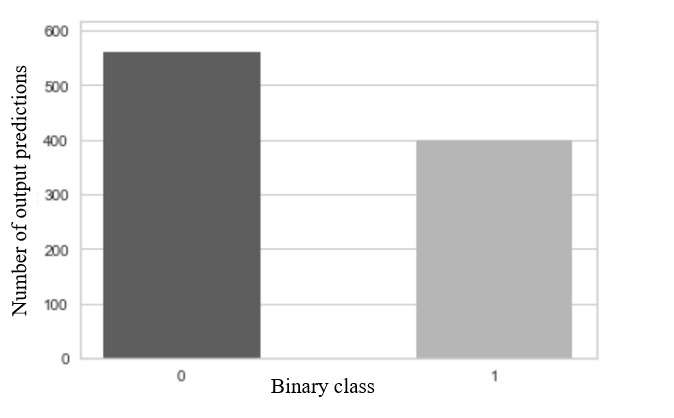}
\caption{Bar graphs representing an even class distribution ($\approx$ 58/42) of stock up or down in the original dataset. }
\end{figure}

\section{Results}

The initial model configuration, described in the previous section, gives an impressive classification (testing) accuracy of \textbf{74.58\%}. A confusion matrix, displayed in figure 6, summarizes the classification performance of the model. 

\begin{figure}
\centering
\includegraphics[height=2.5in,width=0.5\columnwidth]{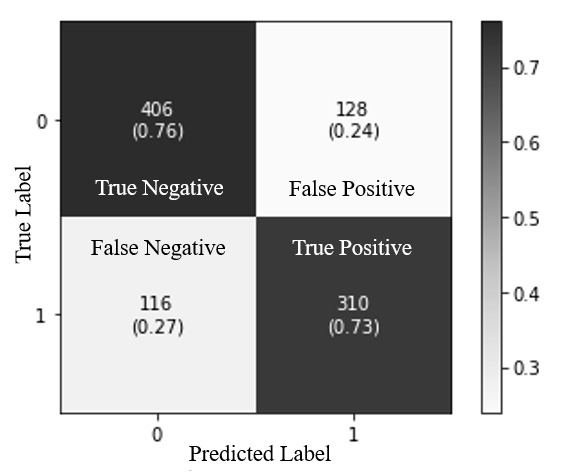}
\caption{Confusion Matrix showing classifier performance at the deep functional level.}
\end{figure}

It indicates that the model is able to correctly identify the 0 class and 1 class with an accuracy of 76\% and 73\% respectively. The F1 score for this configuration is 71.76\%, reinforcing model performance as reflected in the testing accuracy result. The achieved accuracy is far superior to the random guessing threshold of 50\%. This result indicates the effectiveness of NNs in this task, which is in contrast with the results of the correlation analysis performed previously. This output reinforces the claim that NNs have the ability to detect nuanced patterns and produce complex mappings between the input and output. 

\subsection{Effect of Splitting Dataset}
In the original model, the concatenated tweets, resulting in 48 samples, were split into 100 subsets per input sample to augment the dataset. To investigate the effect of modifying the number of subsets per sample on overall performance, values for the input splits were selected in the range [25, 450]. Figure 7 shows the dependence of classification accuracy as a function of split size.

In general, selecting large values of split size leads to performance degradation. Splitting up all tweets on a particular day into higher subsets can result in insufficient information contained within a unique sample, deteriorating prediction capability. It is reasonable to assume that on any given day, some tweets cause the price to go in the opposite direction to that observed in the stock market, however, the aggregate impact of other tweets outweigh this effect. As a result, higher subdivisions do not accurately capture the true nature of the task. To determine if this trend continues, an extreme case was considered i.e. the maximum logical split value was considered. Using all 60,233 filtered tweets as individual inputs to the network, the observed accuracy was 62.37\%, validating the observed graphical results. On the other end of the spectrum, using the concatenated tweets in their unaltered form will reflect all the necessary information on a given day for the model to make predictions. However, in this work, this will entail using a considerably limited number of training instances, hindering the network's ability to learn effectively and leading to erroneous outputs. The training results of this configuration further confirmed this intuition as it gave the worst performance in comparison with using other subset values. The training time for this configuration i.e. using no splits was also significantly higher than any other value tested. As the input sequence length is maximum in this case, a significant number of LSTM cells is required. This discernibly increases processing time and degrades performance due to the emergence of vanishing gradients. Therefore, there exists a trade-off between loss of information and creating a reasonably sized dataset. In light of this fact, a split size of 150 per day was selected for subsequent analyses as it is able to achieve a satisfactory balance of the aforementioned performance variables. 

\begin{figure*}
\centering
\includegraphics[height=2.3in,width=\textwidth]{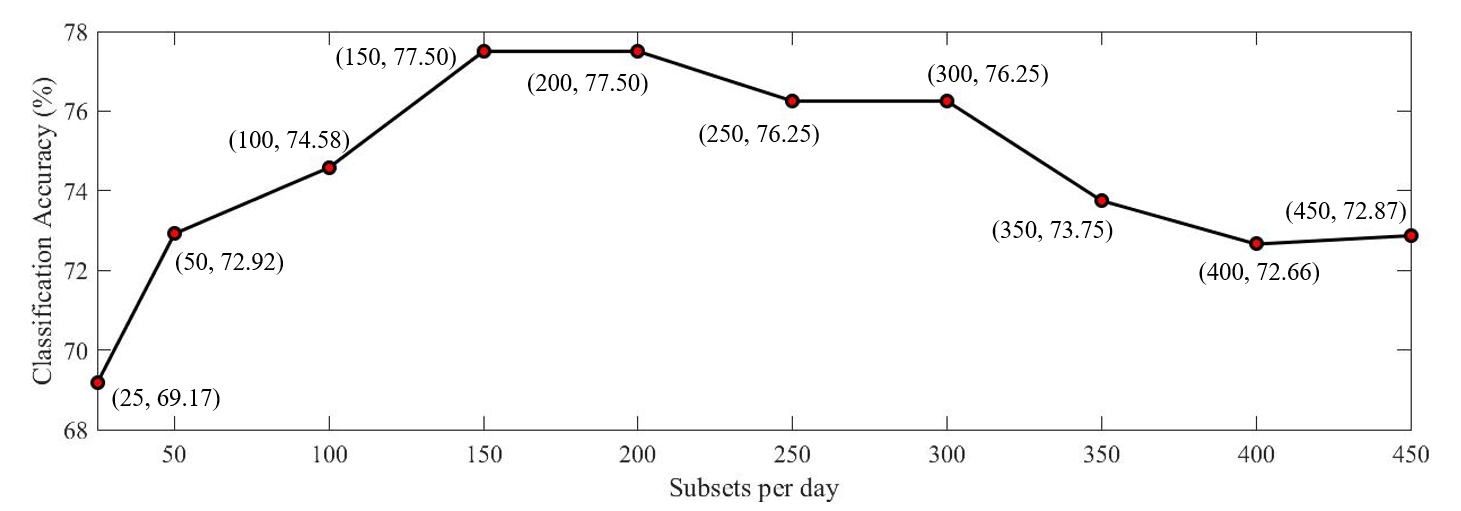}
\caption{Variation of classification accuracy with the split size per input sample.}
\end{figure*}

\subsection{Hyperparameter Optimisation}
To improve the results of the original model configuration, the hyperparameters employed in the NN were optimised. In particular, the dropout rate, batch size, and number of LSTM output hidden units were varied to investigate their impact on classification results. Each hyperparamter was considered in isolation, with all other model variables remaining unchanged, to determine its exclusive impact. The same experimental procedure was also applied to a different network configuration known as bidirectional LSTMs. In strict terms, the standard LSTM structure used in this work is called a unidirectional LSTM network which differs from the bidirectional LSTM structure. The bidirectional network is a variant of the classical LSTM, where information flows in both directions between LSTM cells such that the cell, at every time step, is able to maintain previous and future input information \cite{thirtyone}. This is in contrast to the unidirectional structure, where each cell contains only past information.

Table 1 shows the classification accuracies achieved by varying the network dropout percentage. It is apparent that the unidirectional structure tends to perform well on dropout values higher than 0.2. However, the bidirectional structure shows no notable trends and as such, no conclusive inference can be drawn. The best accuracy (78.12\%) is achieved by the unidirectional architecture, using a dropout value of 0.3. The implications of utilising higher values of dropout are that a higher proportion of neurons become inactive, increasing the robustness of the model and resulting in better testing accuracies.

\begin{table}


  \begin{center}
 
    \caption{Effect of varying dropout rates} 
\begin{tabular}{c c c| c c}
 
  \hline \hline
  \Tstrut
  \multirow{2}{*}{\small{Value}} 
      & \multicolumn{2}{c|}{\small{Unidirectional}} 
          & \multicolumn{2}{c}{\small{Bidirectional}} \\              \cline{2-5}
          \Tstrut
 & \small{Accuracy} & \small{Epoch}  & \small{Accuracy} & \small{Epoch}  \\ \hline
 \Tstrut
      \footnotesize{0.2} & \footnotesize{74.58} & \footnotesize{6,9}& \footnotesize{75.63}& \footnotesize{7} \\

      \footnotesize{0.3} & \footnotesize{\textbf{78.12}} & \footnotesize{4}& \footnotesize{74.79}& \footnotesize{3} \\
      
      \footnotesize{0.4} & \footnotesize{77.29} & \footnotesize{7}& \footnotesize{72.71}& \footnotesize{8} \\
      
      \footnotesize{0.5} & \footnotesize{76.25} & \footnotesize{6}& \footnotesize{75.63}& \footnotesize{3} \\
      
      \footnotesize{0.6} & \footnotesize{75.00} & \footnotesize{5,10}& \footnotesize{\textbf{76.46}}& \footnotesize{7}\\
  \hline \hline   

\end{tabular}
\end{center}
 \end{table}

\begin{table}


  \begin{center}
 
    \caption{Effect of varying batch size} 
\begin{tabular}{c c c| c c}
 
  \hline \hline
  \Tstrut
  \multirow{2}{*}{\small{Value}} 
      & \multicolumn{2}{c|}{\small{Unidirectional}} 
          & \multicolumn{2}{c}{\small{Bidirectional}} \\              \cline{2-5}
          \Tstrut
 & \small{Accuracy} & \small{Epoch}  & \small{Accuracy} & \small{Epoch}  \\ \hline
 \Tstrut
      \footnotesize{8} & \footnotesize{\textbf{78.33}} & \footnotesize{2}& \footnotesize{\textbf{77.92}}& \footnotesize{8} \\

      \footnotesize{16} & \footnotesize{78.12} & \footnotesize{2}& \footnotesize{75.00}& \footnotesize{1} \\
      
      \footnotesize{32} & \footnotesize{74.58} & \footnotesize{6,9}& \footnotesize{75.63}& \footnotesize{7} \\
      
      \footnotesize{64} & \footnotesize{74.79} & \footnotesize{6}& \footnotesize{72.29}& \footnotesize{6} \\
      
      \footnotesize{128} & \footnotesize{73.54} & \footnotesize{8}& \footnotesize{73.12}& \footnotesize{3}\\
  \hline \hline
  
\vspace{-11.5mm}   
\end{tabular}
\end{center}
 \end{table}
 
 \begin{table}


  \begin{center}
 
    \caption{Effect of varying lstm output hidden units} 
    
\begin{tabular}{c c c| c c}

  \hline \hline
  \Tstrut
  \multirow{2}{*}{\small{Value}} 
      & \multicolumn{2}{c|}{\small{Unidirectional}} 
          & \multicolumn{2}{c}{\small{Bidirectional}} \\              \cline{2-5}
          \Tstrut
 & \small{Accuracy} & \small{Epoch}  & \small{Accuracy} & \small{Epoch}  \\ \hline
 \Tstrut
      \footnotesize{100} & \footnotesize{74.58} & \footnotesize{6,9}& \footnotesize{75.63}& \footnotesize{7} \\

      \footnotesize{128} & \footnotesize{74.38} & \footnotesize{6}& \footnotesize{\textbf{77.08}}& \footnotesize{4} \\
      
      \footnotesize{256} & \footnotesize{75.42} & \footnotesize{3}& \footnotesize{75.21}& \footnotesize{10} \\
      
      \footnotesize{512} & \footnotesize{\textbf{76.46}} & \footnotesize{4}& \footnotesize{74.58}& \footnotesize{2} \\

  \hline \hline  
\vspace{-10mm}
\end{tabular}
\end{center}
 \end{table} 
 
 Table 2, which highlights the effect of variable batch sizes, indicates that there is a decline in the accuracy level with increasing batch size for the unidirectional structure. Mini-batch sizes of 8 and 16 yield comparable results, with batch size 8 achieving a remarkable accuracy of 78.33\%. Similarly to the dropout case, there are no identifiable trends for the bidirectional structure, however the lowest batch size (8) performs the best for this variant of LSTM as well. The batch size determines the number of training instances after which the gradients and weights of the network are updated. The impressive performance of the lower batch sizes can be attributed to a more robust convergence as well as the network's ability to circumvent local minima.

The final parameter used for optimisation is the number of hidden state units of the LSTM cells. The hidden state units determine the dimensionality of the output space of the LSTM layer i.e. the dimensionality of the LSTM output vectors $\textbf{H} = \{h_1,h_2,h_3,...,h_n\}$. As presented in Table 3, the accuracy shows an overall increase with an increase in the output dimensionality for the unidirectional framework. Although performance gains are observed upon using larger values of hidden state size, it is at the cost of an exponential increase in the number of parameters and processing time. Values greater than 512 are not used in this study as the resulting models will be prone to overfitting owing to their marked complexity \cite{thirtyseven}. The bidirectional variant performs better when 128 hidden state units are used however results in a performance degradation for higher values. Due to the inherent characteristics of the bidirectional model, the dimensionality of the LSTM output is double that of its unidirectional counterpart. This ultimately leads to an inordinately complex model that is more prone to overfitting.

Some neural network structures are known to achieve satisfactory results by employing two hidden layers. Therefore, two identical LSTM stacked layers with the same hyperparameter values as the original configuration were integrated in the network. The classification accuracy achieved by this configuration was 75.10\%, which is lower than the value (76.67\%) achieved using a single hidden layer implementation. Hence, it is deemed apposite to forego such an architecture. A notable observation, based on the hyperparameter tuning results, is the performance of the bidirectional structure. It has the ability to preserve past and future values in each cell, thereby allowing the network to gain a fuller context of the information present in the input tweets. In theory, this should lead to improvements in the overall predictions made by the model. However, not only does the bidirectional implementation produce comparable results overall but it also occasionally generates lower accuracies than the unidirectional model.

\subsection{Optimal Model}
 In order to discover the optimal model configuration, experiments were conducted using an exhaustive set of combinations of the hyperparameters. Different combinations of the dropout rate, mini-batch size, and hidden state size were deployed, with the range of hyperparameter values limited to the those outlined in Table 1, Table 2, and Table 3. The combination 0.4 (dropout), 8 (batch size), and 100 (hidden units) produces the best results, obtaining a validation accuracy of 81.04\%. Improvements in accuracy cannot be attained by merely using those hyperparameter values which give the highest accuracy when altered independently i.e. using the combination 0.3,8,512. Therefore, it can be argued that there exists some degree of interaction between the variables when varied simultaneously. 

The optimal model configuration is thus given by the combination 0.4,8,100. To perform an unbiased evaluation of the model, the testing data (960 samples) was used. The model produces a testing accuracy of \textbf{76.14\%} when presented with the unseen testing data. Although the model exhibits a remarkable performance in absolute terms, its results are specific to APPLE and are not generalisable to other technology companies. There could be different patterns and inherent complexities within the twitter datasets of other companies which could lead to similar or contrasting results to that observed in this analysis.
\section{Conclusion}
The objective of this work is to develop a model capable of predicting the direction of next-day stock market fluctuations using twitter messages. Tweets associated with APPLE, regarded among the Big Four technology companies, is used as the basis for this analysis. The primary focus of this work is in the development, configuration, and deployment of an LSTM structure. A correlation analysis is briefly explored to determine the relationship between VADER scores, quantifying the sentiment of tweets, and stock market movement.

Using Point biserial correlation coefficient as the measurement metric, a low correlation value of 0.0812 is obtained. Alternate configurations are considered based on the time taken for information contained in tweets to manifest in market movement and a retweet-weighted average configuration. A delay size of 4 days results in the highest correlation value (0.3037). However, it is apparent from these results that VADER is not able to extract any strong relations between societal sentiment and market direction.

In order to initialise the weights of the LSTM network, GloVe embeddings, pre-trained on a sizeable twitter corpus, are utilised. Due to the limited number of training samples, the effect of splitting the dataset for augmentation is analysed. A value of 150 is selected for splitting the input sequence for each day into subsets as it provides a satisfactory trade-off between information loss and a suitable representation of dataset size. Hyperparameter tuning is performed using the validation set and independently varying the dropout rate, batch size, and hidden unit size to further optimise model performance. To determine the optimum configuration, an exhaustive set of varying combinations of selected parameters is tested. A combination of 0.4,8,100 performs the best on the validation set, achieving a testing accuracy of \textbf{76.14\%}.

Despite the level of accuracy being an impressive standalone result, twitter datasets from other technological companies need to be analysed and contrasted with results from this study. This relative comparison will allow the LSTM network performance to be gauged more accurately and in a broader context, enabling the formation of generalisable results. The application of technical indicators such as historical price data can also be explored in conjunction with the components of Fundamental Analysis used in this task to provide more input information vital to classification.

Word embeddings are effective in projecting words/features occurring in similar contexts within a neighbouring vector space. However, it is common for tweets to contain words expressing opposite sentiments that are collocated. This leads to erroneous representations of these fundamentally different words as similar vectors, mitigating their discriminative ability as required for classification \cite{thirtyeight}. Further work should be undertaken to incorporate linguistic lexicons such as SentiNet to capture the effect of word similarity and sentiment \cite{twentyseven}. An alternative approach is to use SSWE (Sentiment-Specific Word Embeddings) which injects sentiment information into the loss function of neural networks \cite{twentyseven}. This could potentially boost performance though the enhancement of the quality of word vectors.

\bibliographystyle{unsrt}
\bibliography{references}

\begin{thebibliography}{10}

\bibitem{one}
Narendra~Babu Anuradha~Yenkikar, Manish~Bali.
\newblock Emp-sa: Ensemble model based market prediction using sentiment
  analysis.
\newblock {\em International Journal of Recent Technology and Engineering
  (IJRTE)}, 8(2), 2019.

\bibitem{two}
Selena Larson.
\newblock Welcome to a world with 280-character tweets.
\newblock
  \url{https://money.cnn.com/2017/11/07/technology/twitter-280-character-limit/index.html}.
\newblock Accessed: 07 November 2019.

\bibitem{three}
Twitter.
\newblock How to tweet.
\newblock \url{https://help.twitter.com/en/using-twitter/how-to-tweet}.
\newblock Accessed 03 March 2020.

\bibitem{four}
Alexander Pak and Patrick Paroubek.
\newblock Twitter as a corpus for sentiment analysis and opinion mining.
\newblock In {\em Proceedings of LREC}, volume~10, January 2010.

\bibitem{five}
N.B. GKumar and S.~Mohapatra.
\newblock {\em The Use of Technical and Fundamental Analysis in the Stock
  Market in Emerging and Developed Economies}, chapter Introduction.
\newblock Emerald Group Publishing Limited, 2015.

\bibitem{six}
Eugene~F. Fama.
\newblock The behavior of stock-market prices.
\newblock {\em The Journal of Business}, 38(1):34--105, 1965.

\bibitem{seven}
John~R. Nofsinger.
\newblock Social mood and financial economics.
\newblock {\em Journal of Behavioral Finance}, 6(3):144--160, 2005.

\bibitem{eight}
John Kordonis, Symeon Symeonidis, and Avi Arampatzis.
\newblock Stock price forecasting via sentiment analysis on twitter.
\newblock In {\em Proceedings of the 20th Pan-Hellenic Conference on
  Informatics}, PCI ’16, New York, NY, USA, 2016. Association for Computing
  Machinery.

\bibitem{nine}
V.~S. {Pagolu}, K.~N. {Reddy}, G.~{Panda}, and B.~{Majhi}.
\newblock Sentiment analysis of twitter data for predicting stock market
  movements.
\newblock In {\em 2016 International Conference on Signal Processing,
  Communication, Power and Embedded System (SCOPES)}, pages 1345--1350, 2016.

\bibitem{ten}
V.~{Kalyanaraman}, S.~{Kazi}, R.~{Tondulkar}, and S.~{Oswal}.
\newblock Sentiment analysis on news articles for stocks.
\newblock In {\em 2014 8th Asia Modelling Symposium}, pages 10--15, 2014.

\bibitem{eleven}
Ayman Khedr, S.E.Salama, and Nagwa Yaseen.
\newblock Predicting stock market behavior using data mining technique and news
  sentiment analysis.
\newblock {\em International Journal of Intelligent Systems and Applications},
  9:22--30, July 2017.

\bibitem{twelve}
Johan Bollen, Huina Mao, and Xiao-Jun Zeng.
\newblock Twitter mood predicts the stock market.
\newblock {\em Journal of Computational Science}, 2, October 2010.

\bibitem{thirteen}
Franco Valencia, Alfonso Gómez-Espinosa, and Benjamin Valdes.
\newblock Price movement prediction of cryptocurrencies using sentiment
  analysis and machine learning.
\newblock {\em Entropy}, 21:1--12, June 2019.

\bibitem{fourteen}
Zhigang Jin, Yang Yang, and Yuhong Liu.
\newblock Stock closing price prediction based on sentiment analysis and lstm.
\newblock {\em Neural Computing and Applications}, September 2019.

\bibitem{fifteen}
Xiao Ding, Yue Zhang, Ting Liu, and Junwen Duan.
\newblock Using structured events to predict stock price movement: An empirical
  investigation.
\newblock In {\em Proceedings of the 2014 Conference on Empirical Methods in
  Natural Language Processing ({EMNLP})}, pages 1415--1425, Doha, Qatar,
  October 2014. Association for Computational Linguistics.

\bibitem{sixteen}
Kyoung jae Kim, Kichun Lee, and Hyunchul Ahn.
\newblock Predicting corporate financial sustainability using novel business
  analytics.
\newblock {\em Sustainability}, 11(1):1--17, December 2018.

\bibitem{seventeen}
Evita Stenqvist and Jacob L{\"o}nn{\"o}.
\newblock Predicting bitcoin price fluctuation with twitter sentiment analysis.
\newblock Master's thesis, School of Computer Science and Communication, 2017.

\bibitem{eighteen}
Sangeeta Oswal, Ravikumar Soni, Omkar Narvekar, and Abhijit Pradha.
\newblock Named entity recognition and aspect based sentiment analysis.
\newblock {\em International Journal of Computer Applications}, 178(46):18--23,
  September 2019.

\bibitem{nineteen}
Venkateswarlu Bonta, Nandhini Kumaresh, and N.~Janardhan.
\newblock A comprehensive study on lexicon based approaches for sentiment
  analysis.
\newblock {\em Asian Journal of Computer Science and Technology}, 8:1--6, 2019.

\bibitem{twenty}
C.~W. {Park} and D.~R. {Seo}.
\newblock Sentiment analysis of twitter corpus related to artificial
  intelligence assistants.
\newblock In {\em 2018 5th International Conference on Industrial Engineering
  and Applications (ICIEA)}, pages 495--498, April 2018.

\bibitem{twentyone}
C.J. Hutto and Eric Gilbert.
\newblock Vader: A parsimonious rule-based model for sentiment analysis of
  social media text.
\newblock In {\em Proceedings of the 8th International Conference on Weblogs
  and Social Media, ICWSM 2014}, January 2015.

\bibitem{twentytwo}
Diana Kornbrot.
\newblock Point biserial correlation.
\newblock In David~Howell Brian S.~Everitt, editor, {\em The Encyclopedia of
  Statistics in Behavioral Science}, volume~1, page 2352. Wiley, 1 edition,
  October 2005.

\bibitem{twentythree}
Snezana Kustrin and Rosemary Beresford.
\newblock Basic concepts of artificial neural network (ann) modeling and its
  application in pharmaceutical research.
\newblock {\em Journal of pharmaceutical and biomedical analysis}, 22:717--27,
  June 2000.

\bibitem{twentyfour}
Rajalingappaa Shanmugamani.
\newblock {\em Deep Learning for Computer Vision}.
\newblock Packt Publishing Ltd., 2018.

\bibitem{twentyfive}
Aaron~Courville Ian~Goodfellow, Yoshua~Bengio.
\newblock {\em Deep Learning}.
\newblock MIT Press, 2016.

\bibitem{twentysix}
Mahidhar Dwarampudi and N.~V.~Subba Reddy.
\newblock Effects of padding on lstms and cnns.
\newblock {\em ArXiv}, abs/1903.07288, 2019.

\bibitem{twentyseven}
Erion Çano and Maurizio Morisio.
\newblock Word embeddings for sentiment analysis: A comprehensive empirical
  survey.
\newblock {\em ArXiv}, 2019.

\bibitem{twentyeight}
Jeffrey Pennington, Richard Socher, and Christoper Manning.
\newblock Glove: Global vectors for word representation.
\newblock In {\em EMNLP}, volume~14, pages 1532--1543, January 2014.

\bibitem{twentynine}
Sepp Hochreiter and Jürgen Schmidhuber.
\newblock Long short-term memory.
\newblock {\em Neural computation}, 9:1735--80, December 1997.

\bibitem{thirty}
Ralf~C. Staudemeyer and Eric~Rothstein Morris.
\newblock Understanding lstm - a tutorial into long short-term memory recurrent
  neural networks.
\newblock {\em ArXiv}, abs/1909.09586, 2019.

\bibitem{thirtyone}
Klaus Greff, Rupesh Srivastava, Jan Koutník, Bas Steunebrink, and Jürgen
  Schmidhuber.
\newblock Lstm: A search space odyssey.
\newblock {\em IEEE transactions on neural networks and learning systems}, 28,
  March 2015.

\bibitem{thirtytwo}
Serena~Yeung Fei-Fei~Li, Justin~Johnson.
\newblock Recurrent neural network.
\newblock
  \url{http://cs231n.stanford.edu/slides/2017/cs231n_2017_lecture10.pdf}.
\newblock Accessed: 03 March 2020.

\bibitem{thirtythree}
Yaoshiang Ho and Samuel Wookey.
\newblock The real-world-weight cross-entropy loss function: Modeling the costs
  of mislabeling.
\newblock {\em IEEE Access}, PP:1--1, December 2019.

\bibitem{thirtyfour}
Followthehashtag.
\newblock One hundred nasdaq 100 companies – free twitter datasets.
\newblock
  \url{http://followthehashtag.com/datasets/nasdaq-100-companies-free-twitter-dataset/}.
\newblock Accessed: 18 October 2019.

\bibitem{thirtyfive}
Yahoo Finance.
\newblock Apple inc. (aapl), nasdaqgs real-time price. currency in usd.
\newblock \url{https://uk.finance.yahoo.com/quote/AAPL/history?p=AAPL}.
\newblock Accessed: 18 October 2019.

\bibitem{thirtysix}
Gavin Hackeling.
\newblock {\em Mastering Machine Learning with scikit-learn. Apply effective
  learning algorithms to real-world problems using scikit-learn}.
\newblock Packt publishing, August 2014.

\bibitem{thirtyseven}
G.P. Zhang.
\newblock {\em Neural Networks in Business Forecasting}.
\newblock Idea Group Publishing, 2004.

\bibitem{thirtyeight}
Duyu Tang, Furu Wei, Nan Yang, Ming Zhou, Ting Liu, and Bing Qin.
\newblock Learning sentiment-specific word embedding for twitter sentiment
  classification.
\newblock In {\em 52nd Annual Meeting of the Association for Computational
  Linguistics, ACL 2014 - Proceedings of the Conference}, volume~1, pages
  1555--1565, June 2014.

\end{thebibliography}

\end{document}